# Do Facial Expressions Predict Ad Sharing?

# A Large-Scale Observational Study

Daniel McDuff

Microsoft Research, USA

Jonah Berger

Wharton School

University of Pennsylvania, Philadelphia, USA

Daniel McDuff, Microsoft Research, Redmond, Washington; Jonah Berger, Wharton School, University of Pennsylvania, Philadelphia, Pennsylvania.

Correspondence concerning this article should be addressed to Daniel McDuff, One Memorial Drive, Cambridge, MA, 02142. E-mail: damcduff@microsoft.com

**Abstract**

People often share news and information with their social connections, but why do some advertisements get shared more than others? A large-scale test examines whether facial responses predict sharing. Facial expressions play a key role in emotional expression. Using scalable automated facial coding algorithms, we quantify the facial expressions of thousands of individuals in response to hundreds of advertisements. Results suggest that not all emotions expressed during viewing increase sharing, and that the relationship between emotion and transmission is more complex than mere valence alone. Facial actions linked to positive emotions (i.e., smiles) were associated with increased sharing. But while some actions associated with negative emotion (e.g., lip depressor, associated with sadness) were linked to decreased sharing, others (i.e., nose wrinkles, associated with disgust) were linked to increased sharing. The ability to quickly collect facial responses at scale in peoples' natural environment has important implications for marketers and opens up a range of avenues for further research.

Social sharing has a huge impact on consumer behavior.  People often share news, information, videos, reviews, and tweets with their peers, and such word of mouth has a significant impact on what people like, think, and buy.  Indeed, from books and restaurants to movies, music, and even new pharmaceutical drugs, social sharing has a significant and sizable impact on purchase behavior (Chevalier and Mayzlin, 2006; Chintagunta, Gopinath, & Venkataraman 2010; Godes and Mayzlin 2004; 2009; Iyengar, Van den Bulte, & Valente, 2010; Moe and Trusov 2011, see Berger 2014 and Babić Rosario, Sotgiu, Valck, and Bijmol 2016 for recent reviews).

Consistent with this impact, more and more marketers have begun to try to harness the power of social sharing.  Rather than simply using traditional advertising, or paid media, to communicate their message, companies are investing in earned media, hoping to get consumers to share their ads and talk about their brands.  The value of earned media is clear.  Rather than having to pay for each exposure, as in traditional advertising campaigns, in the case of earned media, each additional exposure is usually free.  While there is often an upfront cost to create the content, or potentially seed that content in the appropriate places, if consumers share the content, these additional exposures are costless.  As a result, earned media can be a cheaper and more effective way to spread the word.

The challenge, however, is getting consumers to share. While some online content goes viral, getting millions of shares, other content languishes, barely getting shared at all.  Why are some ads shared more than others?  And can companies better predict which ads will be shared?

We examine this question in the context of online video ads. Video is already over 70 percent of Internet traffic and is growing each year[1]. Facebook users watch over 8 Billion videos each day[2] and Snapchat users watch over 10 billion[3]. The average user spends over 16 minutes watching video ads each month and is exposed to over 30 videos during this time.

In particular, we examine whether facial expressions predict sharing. The face plays a key role in emotional expression (Ekman & Rosenberg, 1997; Niedenthal & Brauer, 2012). People who are happy, for example, have markedly different facial expression from those that are sad. More importantly, facial analysis provides an unobtrusive way of measuring emotional reactions that can be performed quickly, easily, and cheaply at scale. We designed a custom facial action detection algorithm and demonstrate that it can reliably predict emotional expression. Then, using this algorithm, we quantify the facial expressions of thousands of individuals in response to hundreds of video advertisements. Finally, we test whether these expressions predict sharing, and which emotional expressions are most predictive.

This research makes four key contributions. First, while some research has begun to look at the relationship between facial expressions and engagement or purchase (Teixeira, Picard, & Kaliouby, 2014; Teixeira, Wedel & Pieters, 2012) no one has looked at how facial expressions relate to sharing.

Second, the little work that has been done on facial expressions has only examined one or two emotions at a time. Further, it has almost exclusively looked at positive emotions. Thus, we know relatively little about whether negative emotions might increase the sharing of advertisements or inhibit it. By simultaneously investigating multiple emotions, both positive and negative, at once, we can more precisely identify the effect of each of these emotions on what people pass on.

---

[1] http://www.cisco.com/c/en/us/solutions/collateral/service-provider/visual-networking-index-vni/complete-white-paper-c11-481360.html
[2] http://mediakix.com/2016/08/facebook-video-statistics-everyone-needs-know/
[3] http://www.theverge.com/2016/4/28/11526294/snapchat-10-billion-daily-video-views-users-stories

Third, most prior approaches to facial coding have required participants to come into a laboratory and sit in front of a computer loaded with special hardware and software. This is somewhat restrictive, unnatural, and costly and limits the breadth of data that can be collected. In contrast, our approach requires no specialized hardware or software, just a webcam. Participants simply clicked on a link and opted-in through a browser based survey.  Capturing responses in a regular, everyday setting increases ecological validity (Derbaix, 1995) and in-situ studies are valuable as replication of finding previously identified in laboratory setting, increasing the confidence in research.

Finally, our approach is high scalable. Studies of human behavior and emotion such as ours are almost always limited due to the prohibitive cost of manual video coding by expert human coders. Automated coding, however, enables collection of a large amount of data from participants across a range of countries in a quick, and cost-effective manner. Given that expressions of emotions in response to mundane media content can be sparse and that large interpersonal variability exists in non-verbal behaviors it is important to sample a broad number of people. Performing this type of analysis at scale is non-trivial.  Beyond presenting novel results, our work shows that these automated techniques are valid methods for scientific research.

## Social Sharing

Researchers have recently become more and more interested in what drives people to talk and share (see Berger 2014 for a review). Some work, for example, suggests that impression management shapes what gets passed along (e.g., Packard and Wooten 2013; Wojnikci and Godes 2008). People care about self-presentation, or how they look to others, so the better something makes them look, the more likely they are to share it. Consistent with this notion, people are more likely to share their consumption experiences if they go well, rather than badly (De Angelis, Bonezzi, Peluso, Rucker, and Costabile 2012) and are more likely to share news articles (Berger and Milkman 2012) or brands (Berger and Schwartz 2011) that are more interesting.

Though it has not been applied to advertisements, some work also suggests that emotion may increase sharing. When people experience emotions, they often turn to others to regulate those emotions and help them make sense of what they are feeling (Rime 2009). Talking with others can help people understand what they feel and why (Rimé, Mesquita, Philippot, & Boca, 1991; Rosnow, 1980) and can help them relive positive experiences and receive social support around negative ones. Consistent with this perspective, people report greater willingness to share urban legends they feel evoke more emotion (Heath, Bell, and Sternberg 2001) and news articles that contain more emotional words are more likely to make the most emailed list (Berger and Milkman, 2012).

Properly measuring emotion, however, is often challenging. First, while one can ask someone how they feel, self-reports are often inaccurate. People do not always have the best insight into their own emotional states. Someone may report feeling negative, for example, without knowing whether they feel angry or anxious. Because they rely on cognitive elaborations of experienced emotions, self-reports also have trouble picking up quick moment-to-moment emotional shifts over time. Further, the mere act of introspection can alter emotional experience (Kassam & Mendes, 2013).  Reporting, or having to report how one is feeling can affect one's physiological state and thus bias responses.  Facial coding provides a method of objectively capturing expressions in the moment.  Facial coding experts are trained and can view videos offline allowing them to identify subtle responses to a stimulus. Facial actions do not map perfectly to felt emotions; however, in a specific context with a high prior on a set of a few emotions facial coding can be used as a proxy for emotional experiences.

Second, there are methods for attempting to extract emotion from the stimulus itself and while is possible from text it is less feasible with video.  Recent advances in natural language processing and sentiment scoring have allowed researchers and practitioners to estimate the valence, and emotionality, of text. Linguistic Inquiry and Word Count (Pennebaker, Mehl, & Niederhoffer, 2003) for example, estimates positivity and negativity by counting the number of positive and negative words in each document.  But while advances in computer vision have begun to allow automated recognition of images, inferring likely emotional reactions to advertisements is far more challenging.  Systems may be able to recognize that an advertisement contains a dog, for instance, but that, by itself, is not enough information to determine what emotions viewers will feel when watching that advertisement.  Furthermore, stimulus analysis assumes that all individuals will respond in the same way to an advertisement and does not account for the individual differences in how an ad will be appraised.

**Facial Responses**

To address these challenges in measuring emotion, we use facial responses. The face plays a key role in emotional expression (Ekman & Rosenberg, 1997; Niedenthal & Brauer, 2012). Emotions are often hard to measure, but starting with Charles Darwin (1873), researchers have recognized the value of using the face as an outward manifestation or signal of these often otherwise internal states. Since then, decades of quantitative research have revealed reliable patterns in the ways that emotions are expressed on the face. The Facial Action Coding System (FACS), for example, has found consistent facial behaviors associated with anger, fear, joy, surprise, pain and deceit (Ekman & Rosenberg, 1997). Further, of the signals that communicate affective information (physiology, non-verbal behaviors, brain activity) the face is one of the more easily interpreted.

As a result, facial analysis provides an unobtrusive method of passively measuring behavior in-the-moment, with a set of "basic" emotional states linked to prototypic expressions (Ekman et al., 1987). People experiencing joy, for example, tend to make different facial expressions than people experiencing disgust. These different expressions involve different movements of individual facial muscles, that can be coded as specific Action Units (AUs). There is not a one-to-one mapping between single AUs, or groups of AUs, and emotion expressions that transcends all contexts (Bettadapura, 2012; Del Líbano et al., 2018); however, in certain controlled situations they do provide very useful signals about the types of emotions being expressed. Consistent with the value of this approach, marketing scientists have recently started to use facial coding as a measurement tool. Researchers have examined the relationship between smiling and purchase intent (Teixeira, Picard, & Kaliouby, 2014), for example, and the link between positive emotion and engagement (Teixeira, Wedel & Pieters, 2012).

Examining facial expressions avoids some common challenges in studying expressions of emotion. Rather than relying on self-report, recording facial responses to advertisements allows an unobtrusive measurement of expressed emotion response that is not disturbed by cognitive elaboration or reflection. It is difficult for an individual to introspect on their own expressions of emotion and report these in real-time. Rather than simply relying on aggregate measures (e.g., one ad is more positive than another) coding facial responses allows us to examine variation in individual responses (i.e., whether one person reacts more positively than another to the same ad) and whether such variation is linked to sharing. Finally, the advent of automated systems that can accurately detect facial muscle movements (De la Torre & Cohn, 2011) allows for increased scalability, repeatability, and observation in naturalistic environments. Rather than examining a small set of laboratory stimuli, we can investigate how thousands of people naturally react to hundreds of advertisements and code these reactions in a consistent and reliable manner. Rather than studying a single emotion in isolation, as in most prior work, this approach allows us to examine how multiple emotions, experienced at various points in watching an advertisement, might impact sharing.

### The Current Research

We measure a range of facial actions, representing both positive and negative emotions, to examine (1) whether facial actions predict sharing and (2) which actions are most predictive.

There are three main ways emotions might relate to sharing. One possibility is that any emotion increases transmission. As noted previously, people often share emotional experiences with others. In fact, psychological research on the social sharing of emotion argues that 90% of emotional experiences are passed on (see Rime 2009 for a review). Consistent with the notion that emotionality increases sharing, movies and news articles that evoke more emotion are more likely to be shared (Berger & Milkman, 2012; Luminet, Bouts, Delie, Manstead, & Rime, 2000). If all emotions increase sharing, then videos that evoke any facial expressions should be shared more.

Another possibility is that emotional valence drives sharing. The simplest way to characterize emotions is by their valence, or positivity. Some emotions (e.g., happiness) are more positive, while others (e.g., anger) are more negative. Given that impression management shapes what people share, one could argue that people might avoid sharing negative things to avoid communicating negative identities (Tesser & Rosen 1975, see Berger 2014 for a review). People prefer interacting with positive rather than negative others (Kamins, Folkes,& Perner, 1997), so consumers may share positive things to avoid seeming like a negative person or a "Debbie Downer". Similarly, most people would prefer to put others in a good mood rather than a bad one. Indeed, research finds that positive news is more likely to be shared than negative news (Berger & Milkman 2012; Tesser & Rosen 1975). Overall, then, these perspectives suggest that content which evokes facial expressions linked to positive, rather than negative, emotions may be shared more.

A third, more complex possibility, is that different specific emotions have different effects on sharing. In addition to valence, specific emotions differ on a variety of other dimensions (e.g., arousal or certainty, Lerner and Keltner 2000; Smith and Ellsworth 1985). This approach would suggest that even though both disgust and sadness are negative emotions, they may have different effects on whether something is shared.

To test these possibilities, we use a new data collection framework, examining naturalistic facial responses (i.e., viewers in their home environment) of thousands of people watching hundreds of videos. Computer algorithms (Senechal, McDuff and Kaliouby, 2015), built specifically for this task, automatically code of facial responses. These algorithms leverage significant advances in computer science, specifically in the field of machine learning, that have enabled the accurate measurement of subtle facial expression in-situ. Given the very challenging nature of coding FACS AUs computer algorithms are still not capable of accurately coding all actions. Therefore, we selected five AUs which are typically associated with expression of different emotions and for which the algorithms had a high degree of accuracy, precision and recall. We examine whether facial expressions reliably predict sharing and, if so, which of five facial actions are most predictive. Our study is the largest of its kind using an online framework and the first to investigate the relationship between emotional responses to video ads and sharing.

We examine both positive and negative emotions, but one might wonder whether advertisements would ever aim to elicit negative emotions. After all, if the goal is to increase consumer evaluations, or encourage purchase, why would a brand ever want to associate themselves with negativity? A few points are worth noting. First, some ads seem to purposely evoke negative emotions. BMW ads, for example, sometimes involve chase scenes or anxiety producing hairpin turns on rocky cliffs. Similarly, paper towel ads may show a disgusting mess to demonstrate how well the paper towels can clean it up. Thus, while positive emotions certainly seem more prevalent, negative emotions do exist in ads and we examine their link to sharing. Second, some ads may induce negative emotion without intending to. One person might find an ad funny while another finds it disgusting or possibly offensive. Someone might find an ad clever while another finds it confusing. Thus, examining individual heterogeneity in facial responses provides valuable insight.

# Method

## Materials

We recorded emotional reactions to 230 video advertisements from a variety of product categories (instant foods, confectionary, pet care, and beauty products). The videos were between 20 and 120 seconds in length (mean duration = 27.3 seconds; SD =8.65 seconds) and from five different countries: Germany (70 ads), US (60 ads), France (40 ads), UK (40 ads), and China (20 ads). Participants viewed ads from their country in their native language. The ads were all recent (aired in the past 10 years) and from major brands. Twenty ads from seven brands were taken from the instant foods category (e.g., Uncle Ben's, Dolmio). The 120 ads from 31 brands were taken from the confectionary category (e.g., Mars, Snickers, Twix, Kinder). The pet foods category consisted of 68 ads from 12 brands (e.g., Pedigree, Cesar, Frolic) and the beauty category consisted of eight ads from seven brands (e.g., Pantene, Triple Velvet, Old Spice). The remaining 14 ads were taken from alcoholic beverage (5), automotive (3) and services (6) ads from 11 brands.

## Participants

Participants (N = 2,106; mean age = 33.6 years; 51% male) were recruited from an online market research panel. From observation of the collected videos, most participants completed the survey from home. The participants were compensated with the equivalent of $8.10 in their local currency. This sample size is an order of magnitude larger than most studies of non-verbal responses to advertisements. However, given the variability in non-verbal behavior between individuals it is important to consider large populations.

**Procedure**

Participants were contacted via email. They were told they were taking part in a study to evaluate video advertisements. Participants simply clicked on a link and opted-in through a browser based survey. They were asked for consent to use their webcam to record while they took part in the study. Participants only needed an Internet connection and a webcam to take part. There was no requirement for specialized hardware or to download or install software. Consequently, participants' experiences while taking the survey were similar to watching online content during everyday life.

The 230 ads were divided into sets of 10. This was done to minimize the effects of boredom. Each participant watched one set of 10 ads and video order within the set was randomized. At the end of each video, participants provided their willingness to share ("If you watched this ad on a website such as YouTube how likely would you be to share it with someone else?" 1 = very unlikely to 5 = very likely). The participants completed the study from home and the average time to finish the survey was 26.3 minutes.

**Automated Facial Coding**

Automatically detecting spontaneous facial actions in everyday environments is challenging. Designing automated facial coding algorithms is difficult, it is not possible to detect every single facial action unit with the requisite performance in everyday settings, due to the subtly and variability in how they appear in videos. Therefore, we designed custom facial action detection algorithms focusing on five of the most commonly occurring, informative and reliably detected actions: smiles, outer eyebrow raises, brow furrows, lip corner depressors and nose wrinkles (see Figure 1 for examples).  Though anxiety is a common negative emotion more generally, it is not a common reaction to advertisements.

**Figure 1. Posed and spontaneous examples of facial actions, and emotions commonly associated with them.**

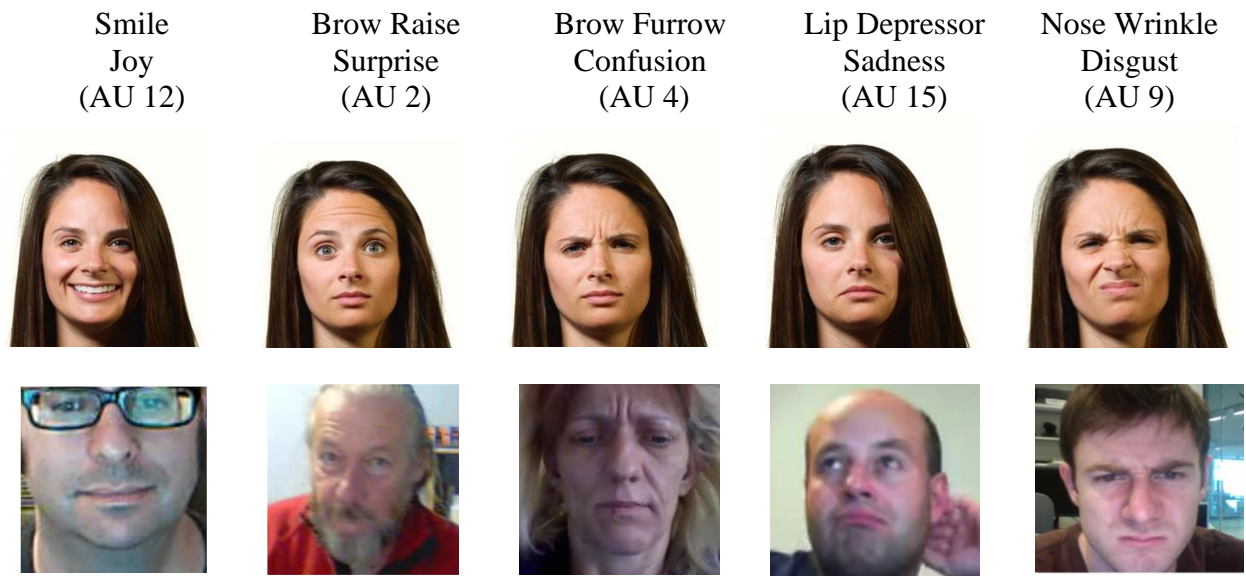

"Smiles" are defined as contractions of the *zygomatic major* muscle, which pulls the lip corners toward the ears (AU 12 in FACS, see Ekman, Friesen, & Hager, 2002 for smiles and other facial expressions). AU 12 is often associated with positive affect. "Outer brow raises" are defined as contractions of the *frontalis (pars lateralis)* muscle, which pulls the eyebrows upward (AU 2 in FACS). AU 2 is often associated with surprise.  "Brow furrows" are defined as contractions of the *corrugator supercilii* muscle, which pulls the eyebrows down and together to form vertical wrinkles on the inner brow (AU 4 in FACS). Researchers commonly interpret AU 4 ("Brow knitting") a signal of mental effort (Oster, 1978) and confusion, worry and concentration are specific affective states associated with it (Rozin and Cohen, 2003). Lip corner depressors" are defined as contractions of *depressor anguli oris* muscle, which pulls the lip corners down (AU 15 in FACS). AU 15 is often associated with sadness. "Nose Wrinkles" are defined as contractions of the *levator labii superioris alaeque nasi* muscle, which pulls the eyebrows down and the nose corners upwards (AU 9 in FACS). AU 9 is often associated with disgust (Kassam, 2010).

**Figure 2. Examples of video frames from our dataset. Data collected in natural environments is more ecologically valid but presents challenges for analysis.**

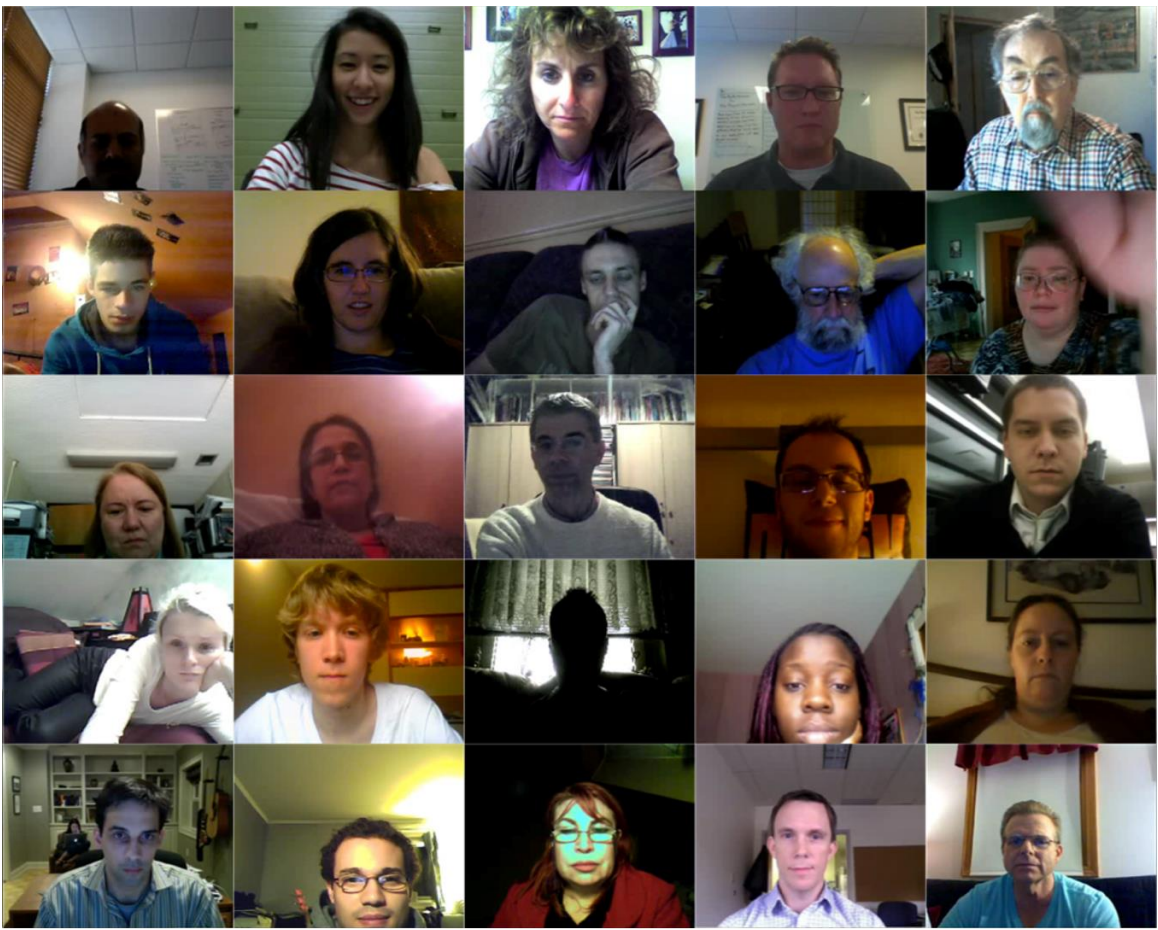

The automated software is designed to detect these facial actions. The software has two principal constituent components, both of which are created using *supervised learning*.

Supervised learning is a machine learning approach that involves training a mathematical model via a set of labeled examples. The first component involves face tracking, it identifies landmarks on the face (e.g., specific locations around the eyes, mouth and nose) using a computer vision technique known as supervised descent. Once the face is located within each video frame, a region of interest is identified including the mouth, nose, eyes and eyebrows. The second component analyzes how the texture of the face region of interest changes, in order to identify which actions are present (e.g. a smile), this component is implemented using a computer vision algorithms known as support vector machines. The output is a probability score for each action. When no expression is present these probabilities are all zero, when one or more actions fire, the corresponding action probabilities will rise. The software computes these probabilities for every frame of video (14 times per second). As people may exhibit and expression for a short time and then return to a neutral expression it will capture these dynamics with a resolution of approximately 70 milliseconds, thus avoiding problems with dial-based methods of moment to moment measurement where participants may forget to turn the dial for several seconds.

Automatic detectors were trained using example videos manually coded by expert human observers. Twenty coders were given training from the Facial Action Coding System (FACS) manual (Ekman, Friesen, & Hager, 2002) and were certified by passing the FACS Test. They were shown similar participant videos collected from a prior study and asked to code the presence of each of the five facial actions as defined in the FACS in each frame of the video. Agreement among expert human coders regarding the presence of a given action was high (Table 1).

Using this expert coded data, the detectors were then trained using *supervised learning* (see McDuff, 2014). The manually coded videos were partitioned into *training* (80,000 labeled video frames), *validation* (10,000 labeled video frames), and *testing* (900,000 labeled video frames) sets. The training set was a random sample of from 4,000 unique individuals, and the validation set was a random sample of from 2,500 unique individuals. Using the training set, we trained a two-class support vector machine (SVM) with a Nystrom-approximated radial basis function (RBF) kernel (Senechal, McDuff & Kaliouby, 2015). Using the validation data, we optimized the number of samples used in the Nystrom approximation ($N_s$ = {200, 500, 1000, 2000}), the SVM cost parameter (C = {0.01, 0.1, …, 100}), and the RBF spread parameter ($\gamma$ = {0.01, 0.1, …, 100}). For more detail about implementation and validation of this approach, see Senechal, McDuff & Kaliouby, 2015. Agreement between machine coding and human coders was high (Table 1, Frame-Level κs > .74) and similar to agreement between humans (Table 1, Frame-Level κs > .55). Other published work has illustrated how the trained models produced using this approach as robust to changes in ethnicity (McDuff, Girard and Kaliouby, 2016), gender (McDuff et al., 2017) and age (McDuff, 2017). The trained classifiers described above are the same as those used in the AFFDEX software development kit (McDuff et al., 2016) which is publicly available. This allows other researchers to use the same system for future experiments. The algorithms also leverages an estimate of each subject's "neutral" face. A moving time window is used to baseline estimates based on the temporal dynamics of normal facial responses.

**Table 1. Inter-rater Coding Statistics. Human-human reliability and human-machine reliability.**

| Action | Human Coding Frame-Level κ | Machine Coding Frame-Level κ |
|---|---|---|
| Smile (AU 12) | 0.81 | 0.74 |
| Brow Raise (AU 2) | 0.75 | 0.88 |
| Brow Furrow (AU 4) | 0.84 | 0.77 |
| Lip Depressor (AU 15) | 0.80 | 0.90 |
| Nose Wrinkle (AU 9) | 0.55 | 0.92 |

Descriptively, the most commonly observed expression was smiles (5.95%), followed by lip depressors (3.38%), eyebrow furrow (3.31%), eyebrow raises (1.80%), and nose wrinkles (0.45%).

**Model**

We use a linear mixed effects model to examine the relationship between facial actions and sharing. This model captures the effect of facial actions on sharing, accounting for the fact that some ads might be more likely to be shared, or that people from certain countries might have a higher propensity to share.

$$\begin{aligned}\text{Sharing} = \quad & \beta_0 + \\ & \beta_1 \text{Smiles} + \beta_2 \text{Eyebrow Raise} + \beta_3 \text{Eyebrow Furrow} + \\ & \beta_4 \text{Lip Depressor} + \beta_5 \text{Nose Wrinkle} + \\ & Z_1 \text{Subject} + Z_2 \text{Ad} + Z_3 \text{Country} + E \end{aligned}$$
(1)

Here, $B_0$ is an intercept, $\beta_1$, $\beta_2$, $\beta_3$, $\beta_4$ and $\beta_5$ are the parameters that estimate the marginal linear effects of Smiles, Eyebrow Raises, Eyebrow Furrows, Lip Depressors and Nose Wrinkles on sharing, $Z_1$, $Z_2$ and $Z_3$ are parameters describing the variance in sharing that can be explained by the differences among Subjects, Ads and Countries, respectively. E is an error term. Modeling subject, ad and country as random effects, means we are not interested in the specific effect of any one subject, ad or country but rather want to account for the overall variability they exert on sharing. This allows us to control for content related factors unrelated to facial expressions that might impact sharing. We tested interactions between the facial actions but did not find a significant difference in the models and therefore, following the principal of Occam's Razor that a simpler solution is better, we discarded interaction terms between facial actions.

In the equation above the score for each of the actions was calculated by first thresholding the classifier output, at the threshold determined in the classifier validation process, and then calculating the fraction of frames in which the output was above the threshold (we can described this as the action "base rate".)

## Results

Results indicate that smiles (AU 12) were positively, and most strongly associated with sharing ($\beta = 1.45$, $SE = 0.03$, $p < 0.001$, Table 2). A 30% increase in smiling is associated with a 10% increase in willingness to share.

Some negative emotions seemed to decrease sharing: Lip depressor (AU 15, often associated with sadness) and brow furrow (AU 4, often associated with confusion) were both negatively associated with sharing ($\beta = -0.17$, $SE = 0.05$, $p < 0.01$ and $\beta = -0.18$, $SE = 0.05$, $p < 0.01$).

Other negative emotions, however, seemed to increase sharing: Nose wrinkles (AU 9, often associated with disgust) were positively associated with sharing ($\beta = 0.22$, $SE = 0.11$, $p < 0.05$).

**Table 2. Link between emotional expressions and sharing. Examples of facial actions, and emotions commonly associated with them. Standard errors shown in brackets.**

| Action | Example | Commonly Associated Emotion | Link to Sharing |
|---|---|---|---|
| Smile (AU 12) | 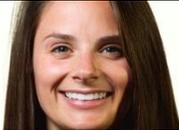 | Joy | 1.45*** (0.03) |
| Nose Wrinkle (AU 9) | 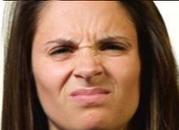 | Disgust | 0.22* (0.11) |
| Lip Depressor (AU 15) | 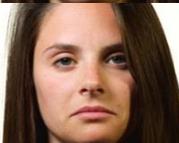 | Sadness | -0.17*** (0.05) |
| Brow Furrow (AU 4) | 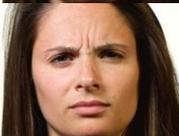 | Confusion | -0.18*** (0.05) |
| Brow Raise (AU 2) | 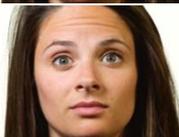 | Surprise | -0.11 (0.07) |

\* $p < 0.05$, \*\* $p < 0.01$, \*\*\* $p < 0.001$

**Effects Over Time**

Given prior work suggesting the time course of emotional experiences affects judgments (e.g., Kahneman, Fredrickson, Schreiber, and Redelmeier, 1993) one might wonder whether experiencing certain emotions early versus late might have different impacts on sharing. To test this, we also built a time varying model. Ads vary in length and structure and as a result modeling time as a continuous variable, even with a large amount of observations is challenging. However, most ads do have some high-level structure. The simplest model to describe this would be a three-act composition of: set-up (beginning), confrontation (middle) and resolution (end). Therefore, we segmented them into thirds (beginning, middle and end) and measured the presence of facial action variables in each segment. A more complex modeling of time could be performed but given the exploratory nature of this work we wanted to use a simpler model of structure. We estimated the following model:

$$\begin{aligned}
\text{Sharing} = \ & \beta_0 \\
& + \beta_{11}\text{Smiles} + \beta_{12}(\text{Smiles}*\text{time}) \\
& + \beta_{21}\text{Eyebrow Raise} + \beta_{22}(\text{Eyebrow Raise}*\text{time}) \\
& + \beta_{31}\text{Eyebrow Furrow} + \beta_{32}(\text{Eyebrow Furrow}*\text{time}) \\
& + \beta_{41}\text{Lip Depressor} + \beta_{42}(\text{Lip Depressor}*\text{time}) \\
& + \beta_{51}\text{Nose Wrinkle} + \beta_{52}(\text{Nose Wrinkle}*\text{time}) \\
& + Z_1\text{Subject} + Z_2\text{Ad} + Z_3\text{Country} + E \quad (2)
\end{aligned}$$

Here, $B_0$ is an intercept, $\beta_{11}$, $\beta_{21}$, $\beta_{31}$, $\beta_{41}$ and $\beta_{51}$ are the parameters that estimate the marginal linear effects of Smiles, Eyebrow Raises, Eyebrow Furrows, Lip Depressors and Nose Wrinkles on sharing. $\beta_{12}$, $\beta_{22}$, $\beta_{32}$, $\beta_{42}$ and $\beta_{52}$ are the parameters that estimate the time dependent effect of each action on sharing. As in the prior model we treat Subject, Ad and Country as random effects. Time is treated as a discrete variable in which the ads were divided into evenly sized temporal bins (N = 3) to capture the effect of expressions during the beginning, middle and end of the ads. As before, we tested interactions between the facial actions but did not find a significant difference in the models.

Results indicate that the relationship between smiling (AU 12) and sharing increased over time ($\beta = 0.05$, $SE = 0.02$, $p < 0.01$). While smiles at the beginning of the video were positively linked to sharing, consistent with the notion the end of emotional experiences has a strong impact (Kahneman, et al, 1993) smiles at the end had an even more positive effect. Though directionally similar, time effects for other actions did not reach significance.

**Alternative Explanations**

One could argue that rather than reflecting sharing, participants sharing responses simply indicated their reactions to the videos. Three things cast doubt on this possibility. First, such an account has difficulty explaining why some negative emotions (i.e., disgust) seem to be positively linked to sharing while others (e.g., sadness) seem to be negatively linked to sharing.

Second, results for liking are different than for sharing (see Table 3). Participants also rated how much they liked each video ("How much did you like the ad you just watched?" 1 = not at all to 5 = very much). Running the same analyses predicting liking rather than sharing shows different effects (Table 3). Facial actions commonly associated with positive emotions increased liking and facial actions commonly associated with negative emotions decreased liking. While nose wrinkles (often associated with disgust) increased sharing, for example, they decreased liking ($\beta = -0.37$, $SE = 0.12$, $p < .05$).

**Table 3: Regression coefficients for time varying liking model (overall base rates) and the difference between the sharing and liking models (standard errors in brackets).**

|  | LIKING Time Varying Model | Difference Sharing vs. Liking |
|---|---|---|
| Intercept | 3.45*** (0.07) | 0.66*** (0.01) |
| Smile | 1.56*** (0.04) | -0.06 (0.04) |
| Eyebrow Furrow | -0.18^ (0.10) | -0.38*** (0.10) |
| Lip Depressor | -0.36*** (0.04) | -0.14 (0.05) |
| Nose Wrinkle | -0.19* (0.05) | 0.06 (0.05) |
| Eyebrow Raise | -0.37* (0.12) | -0.80*** (0.13) |

* p<0.05, ** p<0.01, *** p<<0.01

Third, our main results persist even controlling for liking. The fact that facial expressions have different relationships with liking and sharing cast doubt on the possibility that our sharing measure is merely picking up liking of the videos.

**Role of Culture**

Given that our Internet based framework allowed us to collect data across multiple countries, we were also able to explore cultural variation. Cultures vary in "display rules" or when and where it is acceptable to express certain emotions (Matsumoto, 1990; Matsumoto et al., 2008; McDuff, Girard & el Kaliouby, 2016; McDuff & Girard, 2017; Tsai & Chentsova-Dutton, 2003). In some cultures, expressing strong emotion is supported, but in others it is seen as inappropriate and discouraged (Matsumoto et al., 2008). Building on these ideas, we performed an exploratory analysis about whether the relationship between facial expressions and sharing varied across cultures. We focus on smiles as the prevalence of other emotional expressions are too infrequent for cross-cultural comparisons to be meaningful and most of the prior work studying cultural differences has analyzed smiling (McDuff, Girard & el Kaliouby, 2016; McDuff & Girard, 2017).

Smiles were always positively associated with sharing, but the exact magnitude varied cross-culturally. It is interesting that these results mirror cross-cultural differences in individualism-collectivism (Hofstede, 2001), one of the main dimensions on which cultures vary. The relationship between smiling and sharing was largest in the United States ($\beta_1 = 1.54$) and United Kingdom ($\beta_1 = 1.54$), but smallest in China ($\beta_1 = 0.58$). The United States and United Kingdom have the largest individualism index (Ind. = 91 and 89, respectively), while China has the smallest (20). France ($\beta_1 = 1.50$, Ind. = 71) and Germany ($\beta_1 = 1.29$, Ind. = 67), which have intermediate individualism indices, also showed intermediate relationships between smiling and sharing.

The frequency of smiling (smile base rate) also varied between countries but this does not explain the differences in magnitude of the effects described above. The smile base rate was largest in the US (5.6% of video frames features a smile) and followed by Germany (4.5%), France (4.4%), UK (4.3%) and China (1.5%) respectively.

These results illustrate two points. Firstly, the meanings of smiles in different countries can be different (i.e. not all smiles are created equal), for example the weaker relationship between sharing and smiling in China may point to the fact that more of these smiles are social smiles versus felt amusement. Secondly, the overall propensity for people to smile in different countries will vary and this should be considered as we compare observational data cross-culturally. We cannot ignore the social norms that influence behaviors in different markets.

Individuals in different countries saw different videos, making it difficult to infer too much from these differences, but they suggest cross-cultural differences in emotion and sharing is a valuable direction for future work.

**General Discussion**

A large-scale investigation (i.e., thousands of participants and hundreds of pieces of content) suggests that facial responses predict sharing. While smiles (AU 12) and nose wrinkles (AU 9) were associated with increased sharing, brow furrows (AU 4) and lip depressors (AU 15) were associated with decreased sharing.

These results suggest that not all emotions increase sharing. Further, the fact that some facial actions associated with negative emotions increased sharing (i.e., AU 9, linked to disgust) while other facial actions associated with negative emotions decreased sharing (i.e., AU 4 and AU 15, linked to confusion and sadness) suggests that sharing is driven by more than mere valence alone.

Instead, results are more consistent with specific emotions, and potentially arousal (Berger and Milkman 2012). There is ambiguity around arousal associated with discrete emotion states. However, emotions characterized by high arousal (e.g., joy, Smith & Ellsworth, 1985) seem to be associated with increased sharing and while emotions characterized by low arousal (e.g., sadness) seem to decrease sharing. This suggests that advertisements that evoke high arousal emotions may be shared more.

Ancillary results are also consistent with work on recency or end effects in emotional experiences (Kahneman, et al., 1993), suggesting that latter parts of videos and other content may have more impact on sharing. Future work might examine this in text content, looking at whether articles that evoke high arousal emotions towards the end of the piece are more likely to be shared.

**Implications**

These findings offer an important methodological contribution to researchers interested in studying facial expressions.  A simple, online tool allows researchers to collect facial responses from a range of individuals in their natural environment.  Further the fact that automatic coding is as reliable as individuals' manual coding, means that data can be collected and analyzed at scale.  This opens up a range of avenues for further research.

The findings also have a number of practical implications for managers.  First, the results provide suggestions for designing viral content online. Most ads already try to make people smile, but our findings suggest that certain negative emotions (e.g., disgust) may boost transmission as well.  Thus content creators need not avoid all negative emotions, and, in fact, some negative emotions may help content propagation (also see Berger & Milkman 2012).

Second, the tools used here can be useful in ad design and copy-testing more broadly. Rather than relying on evaluations of the entire advertisement as a whole, marketers can use moment-to-moment analysis of facial expressions to determine which particular components may be working as desired, and which should be replaced.  These tools allow companies to determine whether one character can be dropped, for example, or a certain scene shifted, without having to replace the whole piece of content.  While we have applied these measures to sharing, they can just as easily be applied to evaluation and other outcomes.

Third, even once content has been created, these methods may be useful in determining resource allocation.  Should more resources be put behind seeding and showing advertisement A or advertisement B?  By estimating the likelihood of sharing, facial responses can help determine an ad's likely impact.  Further, it can help determine which advertisements might be better suited for television and which for social media, based on the relative expectation of sharing versus other downstream outcomes.

In conclusion, facial expressions provide a valuable tool to predict and understand consumer behavior.

EMOTIONS AND SHARING OF VIDEO ADS 32Grossi, Mapping expressive differences around the world: The relationship between emotional display rules and individualism versus collectivism, Journal of Cross-Cultural Psychology, vol. 39, no. 1, pp. 55–74, 2008

McDuff, D. J. (2014). Crowdsourcing affective responses for predicting media effectiveness (Doctoral dissertation, Massachusetts Institute of Technology). McDuff, D., Girard, J. M., & el Kaliouby, R. (2016). Large-scale observational evidence of cross-cultural differences in facial behavior. *Journal of Nonverbal Behavior*, 1-19.

McDuff, D., Mahmoud, A., Mavadati, M., Amr, M., Turcot, J., & Kaliouby, R. E. (2016). AFFDEX SDK: a cross-platform real-time multi-face expression recognition toolkit. In Proceedings of the 2016 CHI conference extended abstracts on human factors in computing systems (pp. 3723-3726). ACM.

McDuff, D., Kodra, E., el Kaliouby, R., & LaFrance, M. (2017). A large-scale analysis of sex differences in facial expressions. PloS one, 12(4), e0173942.

McDuff, D. (2017). Smiling from adolescence to old age: A large observational study. In Affective Computing and Intelligent Interaction (ACII), 2017 Seventh International Conference on (pp. 98-104). IEEE.

Girard, J. M., & McDuff, D. (2017). Historical heterogeneity predicts smiling: Evidence from large-scale observational analyses. In 2017 12th IEEE International Conference on automatic face & gesture recognition. IEEE.

Mehl, M. R., Vazire, S., Holleran, S. E., & Clark, C. S. (2010). Eavesdropping on Happiness Well-Being Is Related to Having Less Small Talk and More Substantive Conversations. *Psychological Science*.

Moe, W. W., & Trusov, M. (2011). The value of social dynamics in online product ratings